# An end-user coding-based environment for programming an educational affective robot

Cristina Gena, Claudio Mattutino, Alberto Lillo and Enrico Mosca

[1]*University of Turin, Department of Computer Science, C.so Svizzera 185, 10149 Turin, Italy*

**Abstract**
In this paper we present an open source educational robot, designed both to engage children in an affective and social interaction, and to be programmable also in its social and affective behaviour. Indeed the robot, in addition to classic programming tasks, can also be programmed as a social robot. In addition to movements, the user can make the robot express emotions and make it say things. The robot can also be left in autonomous mode and greeting the user, recognizing biometric user's features and emotion, etc.

## 1. Introduction

Designing and developing methods that allow a robot to perform certain specific tasks, as similar as possible to the human ones, brings productive and practical advantages. Robots work at constant high speeds without the need to stop the process, reducing costs, waste and risks within a company. When it comes to robotics it is easy to associate it to the industrial context, being one of the first sectors that invested and recognized the great potential in reproducing human work, but robotics is so much more. Being an interdisciplinary science, it finds numerous applications in various contexts, from biomedical, to military, from industry to space. In such artificial and sophisticated robots, a psychological and educational approach may bring several benefits to teachers and children in particular. This is what happens in the educational robotics field [1], [2], where man and machine work together, accompanying children towards an engaging, creative, and effective teaching approach. The results achieved by this phenomenon are significant, widespread in primary and secondary schools but in clear growth everywhere. The educational robotics approach is simple and practical and accompanies children, which have a natural predisposition to discover, explore and experiment, in learning through play. Together with educational robotics we find coding, the visual programming approach, typically proposed to children [3]), which stimulates mental processes allowing the children to solve problems of various kinds. Sharing the same educational robotics principles, coding allows children to try their hand at new activities such as programming or deepen the basic concepts of other subjects such as science and mathematics. Both activities favor the introduction to a mental process, union of human thought and computer programming, which translates into the ability to tackle problem solving in an algorithmic way: Computational Thinking [4], an expression increasingly used and applied to any situation, which allows to plan strategies and solutions to possible obstacles. In short, a real skill that should accompany students from the beginning of their school career. Computational thinking therefore deserves to be introduced and cultivated since primary school, not only to bring children closer to a conscious use of technology, but also to develop different levels of abstraction that allow them to deepen logical aspects and deeper structures in any kind of situation. Design, planning, teamwork, reasoning, in-depth analysis, creativity, imagination are some of the ambitions of this goal: knowing how to think like a machine, or rather knowing how to think like who the machine has programmed it.

In this paper we present an open source educational robot, designed both to engage children in an affective and social interaction, and to be programmable by the children also in its social and affective behavior. The paper is organized as follows: Section **??** presents a brief overview of the affective interaction, Section 3 presents the robot and its design, Section 4 describes the robot's hardware and software architecture, and Section 5 concludes the paper.

## 2. Background

During the 1990s, a wave of new research on the role of emotions in different areas such as psychology, neurology and sociology ascertained the vital role of emotions in cognitive processes, previously believed to be an interfer-ence with rational thinking [5]. Above all, such research has challenged the old Cartesian dualistic division that emotional experiences reside only in our mind. Instead, emotions are experienced by our entire body, starting with hormonal changes and nerve signals to tensing or relaxing muscles and facial expressions. Not only that, emotions are also built through the interaction between people, or between people and machines. In the mid-1990s, Picard coined the term Affective Computing [6], a specific branch of artificial intelligence that aims to create machines capable of recognizing and expressing emotions. A computer device with the ability to detect and appropriately respond to the user's emotions and other stimuli could collect clues from a variety of sources. Facial expressions, posture, gestures, speech, strength or rhythm of keystrokes, and changes in temperature of the hand on a mouse can mean changes in the user's emotional state, and these can be detected and interpreted by a computer. via sensors, microphones, cameras and / or software logic. This applicative dimension of emotional expression has been an integral part of our project aimed at realizing an educational, social and affective robot, called Wolly, which we will introduce in the next section.

## 3. The Wolly Robot

At the end of 2017, in our HCI lab we carried out a co-design activity with children aimed at devising an educational robot called Wolly [7]. The main objective of the robot we have devised is acting as an affective peer for children: hence, it has to be able to execute a standard set of commands, compatible with those used in coding, but also to interact both verbally and affectively with students. Currently, Wolly can be controlled by means of a standard visual block environment, Blockly[1], see Fig. 3, which is well known to many children with some experience in coding. However, we also have a simpler set of instructions, see Fig. 2, we tested with children (as described in [8]), so that younger children can use basic commands to control its behavior.

We are now focusing on its role as an educational and affective robot capable of being controlled by coding instructions and at the same time interacting verbally and affectively with children by providing a mirror of emotions: by this last term we mean the the fact that the robot exhibits an affective mirroring behavior, that is, once it recognizes a certain basic emotion in the child it will replicate it by empathy. At the moment Wolly recognizes and expresses mirror emotions through its facial expressions, showing in this way a kind of emotional intelligence, which can be defined as as the capacity to perceive and understand both one's own and other's emo-tions, see [9]. As far as emotion recognition is considered, we have trained a deep neural network written in *Python* with the use of Pytorch library[2] and the Emotic dataset[3], also known as EMOTions In Context, which does not only consider the user's face but also the surrounding context to detect the correct emotion. The robot is also able to recognize biometric features thanks to the use of Face Recognition library by Adam Geitgey[4]. The robot can indeed be left in autonomous mode, in which at the moment it is able to carry out both biometric user's features and emotion recognition, and greeting the user. Regarding this last point, Wolly uses Google Cloud Text-to-Speech to synthesize natural-sounding speech that has been used to greet a person when his face is detected. We are now also working on speech recognition and generation thanks to the integration of Google Cloud Text-to-Speech[5] and Google Speech to Text API[6]. We would also integrate in the near future the dialogue module described in Gena et al. [10] that uses algorithms for recognizing groups of synonyms, which can be easily customized by the user and also allows easy management of the user / robot dialogue in a way similar to Aldebaran's QiChat[7].

In addition to classic programming tasks, the robot may in the future be programmed as a social robot, see Fig. 3. In addition to movements, the user may make the robot express emotions and make it say things, programming in this way its dialogue and its social and affective behavior.

## 4. Hardware and Software Architecture

### 4.1. Hardware architecture

Currently we have developed a second release of the robot (as far as the first release is concerned see [7]). We had the requirement of transforming Wolly into a low cost robot that everyone could build regardless of its electronic and technical skills. The idea is to convert the old Arduino-based robot architecture (see details in [11]) in a all-in-one device managed by a Raspberry Pi, with a sort of WollyOS (Wolly Operating System) easy to download and update, also for non-technical users, as teachers. In more details, the list of the main hardware

---

[1] https://developers.google.com/blockly/

[2] https://pytorch.org/

[3] https://github.com/rkosti/emotic, is a database of images with people in real environments, annotated with their apparent emotions. The images are annotated with an extended list of 26 emotion categories combined with the three common continuous dimensions Valence, Arousal and Dominance

[4] https://github.com/ageitgey/face_ecg

[5] https://cloud.google.com/text-to-speech

[6] https://cloud.google.com/speech-to-text

[7] http://doc.aldebaran.com/2-5/naoqi/interaction/dialog/aldialog$_$syntax_overview.html

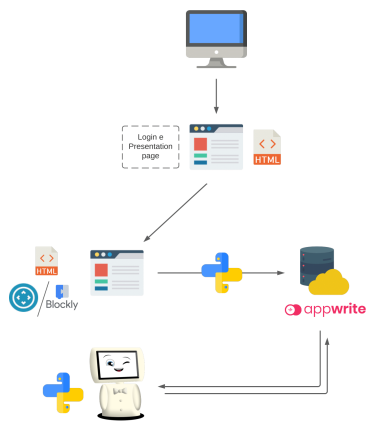

**Figure 1:** Software architecture

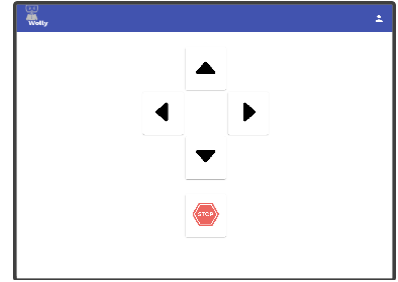

**Figure 2:** The robot can be moved with the 4 arrows in the page, and there is s also a stop button to stop it.

components we are now using is the following:

- Raspberry Pi model 3 or higher
- Screen 5 inch HDMI LCD - for Raspberry Pi 3B + / 4B
- Waveshare Motor Driver Hat for Raspberry Pi - Two DC Motors I2C Interface 5V 2 gear motors 5V
- Tracks Baoblaze in technopolymer/Tamiya 70100
- Bluetooth Speaker JBL Go2
- Raspberry Pi Camera 5MP 1080P
- Powerbank 2000mAh or higher

### 4.2. Software architecture

As far as the software components of the second release are concerned, Wolly is now controlled both from a desktop web-based interface and from a smartphone/tablet application. The responsive web-based application is developed in HTML/CSS and the commands for the robot are sent and read in Python. The mobile application has been developed in Flutter[8] and presents a webview showing the web pages through which the user can take control of Wolly both in iOS and Android. More in detail, we are using the following software components (see also Figure 1).

**Appwrite**[9]. Appwrite is an end-to-end back-end server that is aimed to abstract the complexity of common, complex, and repetitive tasks required for building a modern app. Appwrite provides the developer with a set of APIs, tools, and a management console User Interface (UI) to help developers build apps a lot faster and in a much more secure way. Appwrite offers different services, such as user authentication and account management, user preferences, database and storage persistence, cloud functions, localization, image manipulation, and so on. Appwrite is both cross-platform and technology agnostic, meaning it can run on any operating system, coding language, framework, or platform. Although Appwrite can easily fit the definition of a server-less technology, it's designed to run well in multiple configurations. You can integrate Appwrite directly with your client app, use it behind your custom backend or alongside your custom back-end server.

The communication between Wolly and the other components takes place through the Appwrite database. Every action is written on the database and the robot is always listening for changes (commands). Whenever they occur Wolly will execute them, and when the set of commands will be finished, the database will be reset and ready for receiving a new set of commands.

A basic set of operations we have defined are as follows: (i) The user connects to the website via the mobile or desktop app; (ii) To access the web site she has sign up/ sign in, and all the credentials are stored, managed and checked via Appwrite. At the moment Appwrite is currently running onto a server located in our department; (iii) When the login succeeds, the credentials are retrieved via Appwrite, which then returns an object containing the user-id, which will be used as identifier to store the commands given to the robot into the database; (iv) After the login phase the user may insert a sequence of commands (custom blocks) using the Google Blockly-based interface. When the play button is pressed, all the visual commands are translated in Python, processed and inserted into a list sent to Appwrite. In the near future we will also provide the (more advanced) user with a Python editor; (v) The robot reads the first element of the list and executes the action. When the action is done, the first element is dropped from the list and the cycle re-start until the list is empty.

---

[8] https://flutter.dev/
[9] https://appwrite.io

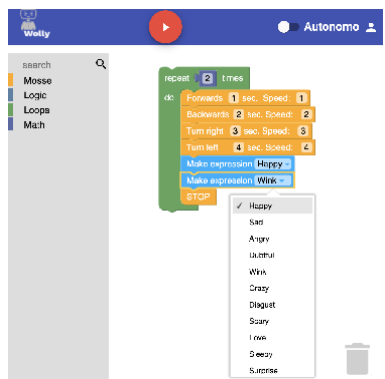

**Figure 3:** The Blockly Page

### 4.3. Blockly

**Blockly**. This library adds an editor representing the coding concepts as interlocking graphical blocks to represent code concepts like variables, logical expressions, loops and, thanks to the customized blocks, to move the robot. As outputs, it returns syntactically correct Python code that is eventually sent to the database.

Blockly is an open-source library, so it's very easy to create and customize your own block.
Via the **Blockly Developer Tools** [10] it is possible to design a special block from scratch by choosing the colors, the shape, how it will be translated and in which language, in our case we have chosen Python.

There are 6 custom designed blocks:

- Move Forward
- Move Right
- Move Left
- Move Backward
- Stop
- Expression

The expression block has a dropdown menu for choosing one of the eleven different available expressions the robot is able to express. .

### 5. Conclusion

In the near future we want to make the robot: be able to carry out a basic vocal interaction through dialogues linked to both the context of the coding exercises and also to its general knowledge about the world and about itself, according to a communication protocol that simulates a natural dialogue.

We are also focusing on the UX simplification. We believe that it is not only important that the user can program the Wolly Raspberry robot with a simple interface, as Google Blockly, but also to easily download its app, and its update, and save it on a micro-SD card (e.g. from the website learn.wolly.di.unito.it). We will also perform an evaluation in the wild, by bringing the robot at school and trying some educational robotics exercises, as well as some end-user programming tasks, such as having the students programming some basic dialogues and interaction withe robots.

---

[10]https://blockly-demo.appspot.com/static/demos/blockfactory/index.html